\documentclass[lettersize,journal]{IEEEtran}
\usepackage{amsmath,amsfonts}
\usepackage{algorithmic}
\usepackage{algorithm}
\usepackage{array}
\usepackage[caption=false,font=normalsize,labelfont=sf,textfont=sf]{subfig}
\usepackage{textcomp}
\usepackage{stfloats}
\usepackage{url}
\usepackage{verbatim}
\usepackage{graphicx}
\usepackage{cite}
\usepackage{multirow}
\usepackage{xcolor}
\usepackage{makecell}
\usepackage{enumitem}
\usepackage{pifont}
\usepackage{bm}

\newcommand\best[1]{\textbf{\textcolor{red}{#1}}}
\newcommand\subbest[1]{\textcolor{blue}{#1}}
\begin{document}

\title{WM-MoE: Weather-aware Multi-scale Mixture-of-Experts for Blind Adverse\\Weather Removal}

\author{Yulin Luo, Rui Zhao, Xiaobao Wei, Jinwei Chen, Yujie Lu, Shenghao Xie, Tianyu Wang, \\ Ruiqin Xiong, Ming Lu, Shanghang Zhang
    \thanks{Y. Luo, R. Zhao, J. Chen, Y. Lu, T. Wang, R. Xiong, and S. Zhang are with Peking University, Beijing, China. 
    (e-mail: \{yulin, ruizhao, 2000012967\}@stu.pku.edu.cn, tianyuw2001@gmail.com, \{cjw, rqxiong, shanghang\}@pku.edu.cn) (Corresponding author: Shanghang Zhang)}
    \thanks{X. Wei is with University of Chinese Academy of Sciences, Beijing, China. (e-mail: weixiaobao0210@gmail.com)}
    \thanks{S. Xie is with Wuhan University, Wuhan, China. (e-mail: xieshenghao@whu.edu.cn)}
    \thanks{M. Lu is with Intel Labs China. Beijing, China. (e-mail: lu199192@gmail.com)}
}

\markboth{Journal of \LaTeX\ Class Files,~Vol.~14, No.~8, August~2021}%
{Shell \MakeLowercase{\textit{et al.}}: A Sample Article Using IEEEtran.cls for IEEE Journals}



\maketitle
\begin{abstract}
Adverse weather removal tasks like deraining, desnowing, and dehazing are usually treated as separate tasks. However, in practical autonomous driving scenarios, the  type, intensity, and mixing degree of weather are unknown, so handling each task separately cannot deal with the complex practical scenarios. In this paper, we study the blind adverse weather removal problem. Mixture-of-Experts (MoE) is a popular model that adopts a learnable gate to route the input to different expert networks. The principle of MoE involves using adaptive networks to process different types of unknown inputs. Therefore, MoE has great potential for blind adverse weather removal. However, the original MoE module is inadequate for coupled multiple weather types and fails to utilize multi-scale features for better performance. To this end, we propose a method called Weather-aware Multi-scale MoE (WM-MoE) based on Transformer for blind weather removal. WM-MoE includes two key designs: WEather-Aware Router (WEAR) and Multi-Scale Experts (MSE). WEAR assigns experts for each image token based on decoupled content and weather features, which enhances the model’s capability to process multiple adverse weathers. To obtain discriminative weather features from images, we propose Weather Guidance Fine-grained Contrastive Learning (WGF-CL), which utilizes weather cluster information to guide the assignment of positive and negative samples for each image token. Since processing different weather types requires different receptive fields, MSE leverages multi-scale features to enhance the spatial relationship modeling capability, facilitating the high-quality restoration of diverse weather types and intensities. Our method achieves state-of-the-art performance in blind adverse weather removal on two public datasets and our dataset. We also demonstrate the advantage of our method on downstream segmentation tasks.
\end{abstract}


\section{Introduction}
In autonomous driving scenarios, adverse weather could severely impact the imaging quality of cameras, leading to the degradation of AI performance~\cite{liu2023deep}. Thus, adverse weather removal is a crucial technique for autonomous driving. Most existing methods handle different weather types separately, such as deraining, desnowing, and dehazing. They can be divided into three categories: task-specific, task-agnostic, and multi-task-in-one methods. The task-specific methods are designed for a specific type of weather, such as deraining~\cite{jiang2020multi, li2018recurrent, ren2019progressive, yu2021single, yang2021end, wang2020deep}, dehazing~\cite{liu2019griddehazenet, dong2020multi, qin2020ffa, wu2021contrastive, li2022usid, lin2022msaff, shin2021region, hu2019adaptive, song2017single}, and desnowing~\cite{liu2018desnownet, chen2020jstasr, chen2021all}. These methods have a weather-related inductive bias, making them difficult to perform well on other tasks. Task-agnostic methods have a unified solution to different tasks but need to be trained separately~\cite{zamir2021multi, liang2021swinir, zamir2022restormer}, and need users to select specific parameters according to the weather type. The applications of task-agnostic methods are quite limited. Hence, their usefulness is limited as well. Multi-task-in-one methods can handle different types of weather using a single set of parameters~\cite{chen2021ipt, valanarasu2022transweather, chen2022unified, kulkarni2022unified}, but they still have several limitations. For example, the schemes of defining weather types in existing methods are complex, and mixed weather types are not considered. BIDeN~\cite{han2022blind} formulates the real-world mixture weather removal as a blind image decomposition task. During training, the multi-domain GAN-based model requires each weather component, which is not available in real scenarios, thus limiting its applicability.

Since the type, intensity, and mixing degree of the weather are unknown in the real world, recent blind weather removal aims to restore corrupted images with unknown weather types. It has gained increasing attention from the community~\cite{li2020all, han2022blind, valanarasu2022transweather, chen2022unified}. The key to blind weather removal is dynamically processing the input based on the weather type. Mixture-of-Experts (MoE)~\cite{han2022survey} is a model that adopts adaptive expert networks to process different inputs with the help of a router. Therefore, MoE has great potential for blind weather removal. Some methods have tried to apply MoE to the weather removal task. HCT-FFN~\cite {chen2023hybrid} utilizes degradation-aware MoE (DaMoE) to extract local features for restoring spatially-varying rain degradation, DRSformer~\cite{chen2023learning} learns enriched sparse content features for deraining via a mixture of experts feature compensator. DAN-Net~\cite{ye2022towards} employs adaptive gated neural to modulate the outputs of task-specific experts, which can be utilized to deal with the mix of snow and haze. However, they neglect using weather features to guide expert selection for the blind weather removal task.

Our work also explores the application of MoE in blind weather removal. We first introduce the MoE module into the ViT-based image restoration network and take it as our baseline. Although performance can be improved by introducing MoE directly, we still find two limitations of this baseline. Firstly, the basic router of MoE cannot well assign the correct weather type to the input due to the coupled weather and content embedding. Secondly, since processing different weather types requires different receptive fields, naive experts fail to exploit the multi-scale features for better performance.

To this end, we propose a method called Weather-aware Multi-scale Mixture-of-Experts (WM-MoE) based on the Transformer for blind adverse weather removal. We use the vision transformer as the baseline, with a task-shared convolution head and tail, and a naive MoE module. To handle diverse adverse weather conditions with multiple weather experts, we design a WEather-Aware Router (WEAR) that can route each image token to specific experts based on decoupled content and weather features. Therefore, WEAR can focus more on weather information to select experts than image content such as texture richness and brightness. To obtain distinctive weather features, we propose Weather Guidance Fine-grained Contrastive Learning (WGF-CL). After obtaining weather tokens by a light-weight ViT encoder, WGF-CL optimizes the mutual distance of these embeddings in a token-level supervised contrastive learning manner, which utilizes the intra-weather similarity and inter-weather difference to guide the selection of positive and negative samples for each token. We further introduce Multi-Scale Experts (MSE) to fuse multi-scale features and enhance the spatial modeling capability, leading to better performance than the original token-wise FFN experts. We demonstrate the effectiveness of the proposed WM-MoE on multiple benchmarks. In the upstream blind weather removal task, WM-MoE surpasses current state-of-the-art (SOTA) methods on the proposed dataset MAWSim by PSNR +1.51 and
SSIM +0.022, the public dataset Allweather~\cite{valanarasu2022transweather} by PSNR +0.96 and SSIM +0.0304, RainCityscapes~\cite{hu2019rain_cityscapes} by PSNR +0.31 and HazeCityscapes~\cite{sakaridis2018haze_cityscapes} by PSNR +1.58. In the downstream task, the image recovered by our method can also improve the performance of the semantic segmentation model, which is more stable than other methods. The contributions can be concluded as follows:
\begin{itemize}
\item We propose a novel framework named Weather-aware Multi-scale Mixture-of-Experts (WM-MoE) for blind adverse weather removal. We design a WEather-Aware Router (WEAR) to assign specific experts for each image token more effectively based on decoupled content and weather features. We also develop Multi-Scale Experts (MSE) to aggregate local and multi-scale features, improving spatial modeling capability.

\item We propose Weather Guidance Fine-grained Contrastive Learning (WGF-CL) to capture discriminative and detailed weather representation. The learned weather features serve as supplementary input for WEAR to select experts, effectively handling complex weather conditions.

\item We conduct comprehensive experiments on two public datasets and our dataset and achieve SOTA blind weather removal performance. We also demonstrate the advantage of our method on downstream segmentation tasks.
\end{itemize}







\section{Related Work}
\subsection{Adverse Weather Removal}
Adverse weather removal has been explored over the past years. Related works can be divided into task-specific and task-agnostic. The adverse weather mainly includes rain, snow, and haze. For task-specific methods, one network aims to deal with certain weather. Li et al.~\cite{li2018recurrent}, Ren et al.~\cite{ren2019progressive}, and Jiang et al.~\cite{jiang2020multi} remove rain based on progressively refining the image. Dong et al.~\cite{dong2020multi} remove haze based on boosting and error feedback. GridDehazeNet~\cite{liu2019griddehazenet} and FFA-Net~\cite{qin2020ffa} use attention operations for dehazing. Wu et al.~\cite{wu2021contrastive} introduce contrastive learning to dehaze. DeSnowNet~\cite{liu2018desnownet}, JSTASR~\cite{chen2020jstasr}, and Chen et al.~\cite{chen2021all} aims to remove snow with different status adaptively.

In contrast to task-specific methods, some methods can be adopted for different weathers, i.e., the task-agnostic methods. Some of these methods can deal with only one type of degradation. These methods need to be trained for each task separately. MPRNet~\cite{zamir2021multi}, SwinIR~\cite{liang2021swinir}, and Restormer~\cite{zamir2022restormer} are architectures for general image restoration. Most of these architectures implement deraining as one of the tasks in the experiments. Some methods can remove multiple adverse weathers at once. All-in-One~\cite{li2020all} uses neural architecture search (NAS) to discriminate between different tasks.
Several strategies are proposed to handle multiple adverse weathers simultaneously. TransWeather~\cite{valanarasu2022transweather} uses learnable weather-type embeddings in the decoder. Chen et al.~\cite{chen2022learning} use a two-stage knowledge-learning mechanism for comprehensive bad weather. BID~\cite{han2022blind} aims to decompose degraded images into constituent underlying images and other components.

\subsection{Transformer in Image Restoration Task}
Transformers~\cite{vaswani2017transformer} have been increasingly applied in the vision area since ViT~\cite{dosovitskiy2020vit} employ Transformers to visual recognition task~\cite{han2022survey}. IPT~\cite{chen2021ipt} introduces Transformers pre-trained on a large dataset for image restoration tasks.
SwinIR~\cite{liang2021swinir} introduces the Transformers with shifted windows~\cite{liu2021swin} for image restoration. UFormer~\cite{wang2022uformer} and Restormer~\cite{zamir2022restormer} use Transformers to construct pyramidal network structures for image restoration based on locally-enhanced windows and channel-wise self-attention, respectively. ELAN~\cite{zhang2022efficient} and DATSR~\cite{cao2022reference} consider long-range attention based on Transformers for super-resolution. CAT~\cite{chen2022cross} and Xiao et al.~\cite{xiao2022stochastic} propose Transformers with adaptive windows to perform more flexible image restoration. Image De-raining Transformer (IDT)~\cite{xiao2022idt} develops the complementary window-based and spatial-based transformer to capture local and non-local features. Dehazeformer~\cite{song2023dehazeformer} explores the limitation of Swin Transformer~\cite{liu2021swin} when applied to haze removal and proposes several improvements including modified normalization layer, activation function, and spatial information aggregation scheme. SnowFormer~\cite{chen2022snowformer} employs cross-attention to model the local-global context interaction across patches for better desnowing. 

\begin{figure*}[!ht]
    \centering
\includegraphics[width=0.85\linewidth]{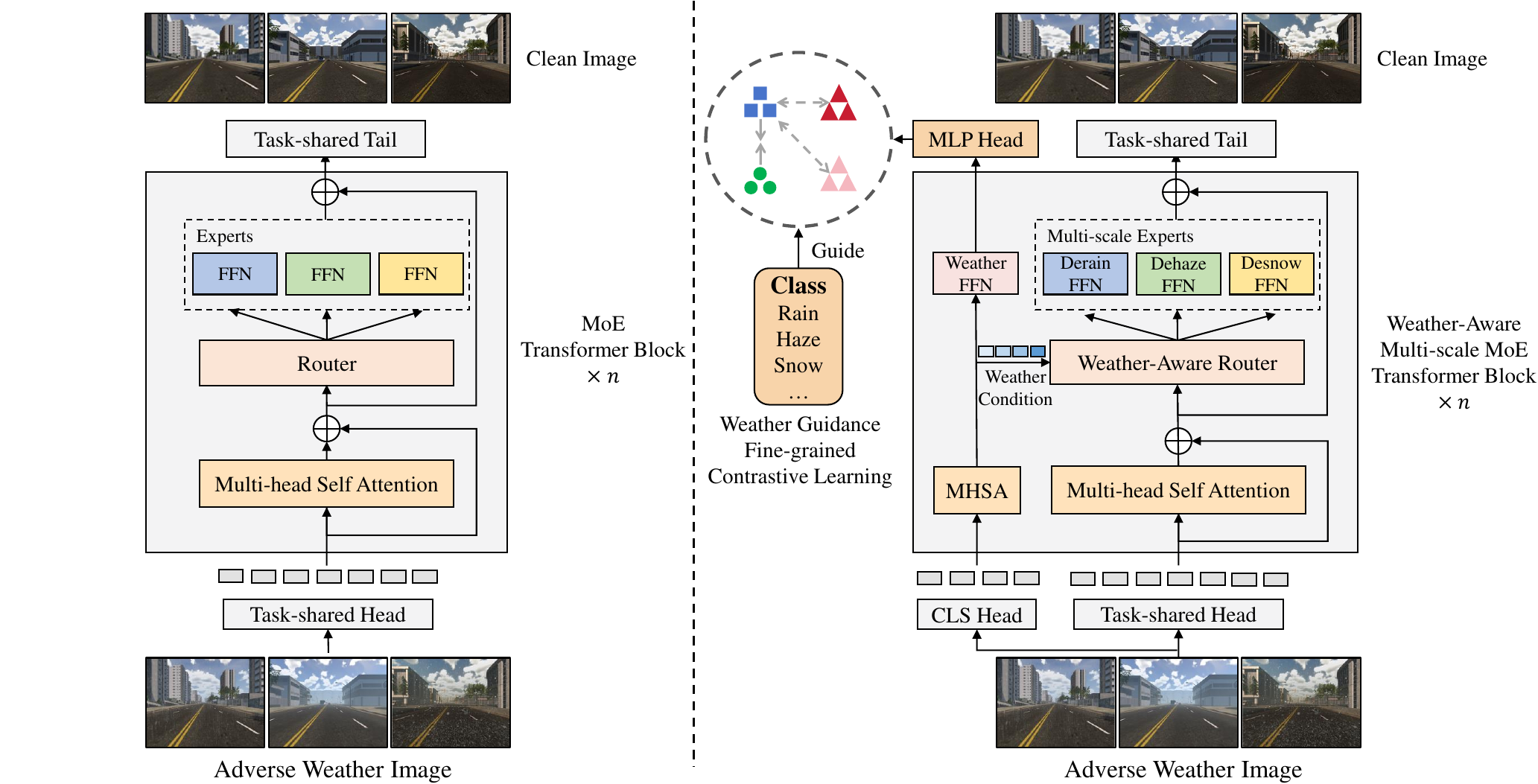}
    \caption{Comparison between (a) naive Mixture-of-Experts (MoE) module~\cite{lepikhin2020moetransformer} and (b) our proposed Weather-aware Multi-scale Mixture-of-Experts (WM-MoE) module. The linear layer router and point-wise FFN experts in the original MoE fail to deal with coupled content and weather features, and local information. Respectfully, we propose a Weather-aware Router to select experts dynamically based on decoupled content and weather embedding and Multi-scale Experts to make full use of local and multi-scale features.}
    \label{fig:method}
\end{figure*}

\subsection{Mixture-of-Experts (MoE)}
MoE is a type of neural network, whose parameters are partitioned into different subsets called "experts"~\cite{fedus2022review}. Different parts of inputs will be routed to specific experts by some router mechanisms in training and inference time~\cite{fedus2022review}. MoE was applied and popularized first in Natural Language Processing (NLP) in the deep learning era. It usually appears as a basic model component
, such as the expert FFN layers in Transformer~\cite{lepikhin2020gshard, lepikhin2020moetransformer}. 

The primary purpose of MoE is to scale up model parameters to achieve better performance while maintaining high computation efficiency~\cite{fedus2022review}. At the same time, the applications of MoE in multi-task learning and cross-domain learning are also explored. MMoE~\cite{ma2018mmoe} utilizes multi-gate routing mechanism to model the relationship of different tasks.~\cite{kudugunta2021beyond}
introduces task-level routing instead of common-used token-level routing to realize more efficient inference. DEMix~\cite{gururangan2021demix} introduces domain experts, each of them specialized in different language domains (e.g. medical, News, etc.) and could be ensembled to better generalize to unseen domains. All these works indicate the potential of MoE beyond scaling up.

In computer vision, MoE has been employed in some high-level tasks such as image classification~\cite{riquelme2021cv-moe-vit, liu2022swinT-v2}, object detection~\cite{liu2022swinT-v2, wu2022residual-moe} and segmentation~\cite{wu2022residual-moe}. MoE has also been used in low-level vision. Literature~\cite{emad2022moesr} and~\cite{liang2022efficient} extract underlying degradation features to construct MoE adaptive network to handle different degradation in blind super-resolution. HCT-FFN~\cite{chen2023hybrid} uses MoE to learn spatially-varying rain distribution features in the deraining task. DRSformer~\cite{chen2023learning} uses MoE to extract sparse content features from rainy images. DAN-Net~\cite{ye2022towards} utilizes adaptive attention gate to modulate the outputs of task-specific experts to handle adverse winter weather conditions. Compared with works mentioned above, we focus on the blind adverse weather removal task. We first explore the limitation of the naive MoE structure when applied to this scenario from token assignment and multi-scale features extraction, and then propose specific MoE design to improve performance.

\section{Proposed Approach -- WM-MoE}
In this section, we first present the overall pipeline of our WM-MoE framework. Then we introduce the baseline, naive MoE module in ViT, and analyze its limitation when applied to the task setting. Afterward, we provide the details of the proposed Weather-aware Multi-scale Mixture-of-Experts (WM-MoE), which includes several key components specially designed for blind weather removal tasks: Weather-aware Router, Weather Guidance Fine-grained Contrastive Learning and Multi-scale Experts.

\subsection{Overall Framework}
The overall pipeline of WM-MoE is shown in Fig.\ref{fig:method} (b). WM-MoE has two parallel branches, one for obtaining weather representation efficiently, and the other for image restoration using task-related features. Both branches are based on Vision Transformer due to their ability in low-level visual tasks~\cite{zamir2022restormer}. 

Given an image $\textbf{I}_{\rm adverse}\in\mathbb{R}^{3\times H \times W}$ with adverse weather, for the weather representation captured branch, we utilize ViT~\cite{dosovitskiy2020vit} to obtain patch-level weather embedding. Then we optimize Weather Guidance Fine-grained Contrastive Learning loss to achieve discriminative features, which is an additional input to Weather-aware Router in the restoration branch.

For the weather removal branch, we first obtain shallow image embeddings by a task-shared convolution head to capture low-level features. After applying patch embedding to image features, tokens are passed through Transformer encoder with $L$ blocks. In each block, self-attention module models global relationships, followed by the proposed Weather-aware Multi-scale MoE module, which includes a Weather-aware Router to assign tokens reasonably and flexibly based on decoupled content and weather embeddings, and Multi-scale Experts to make full use of local and multi-scale information to handle weather with various conditions. After that, we employ a linear layer to expand token dimensions and reshape them to the origin resolution. Finally, a task-shared convolution tail is utilized to refine features and adjust channels to obtain the restoration clean image.

\begin{figure}[!t]
  \centering
  \includegraphics[width=1\linewidth]{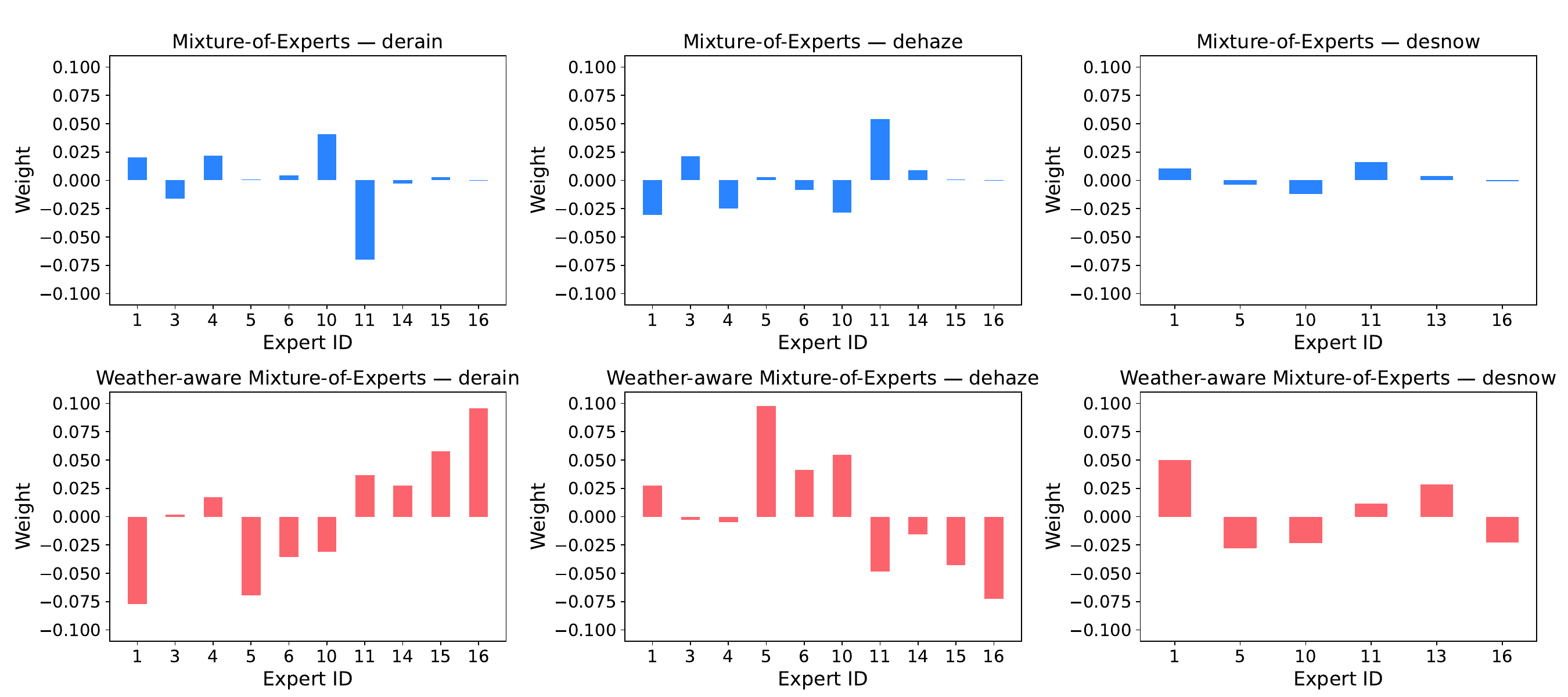}
  \caption{Comparison of normalized routing scores histogram between MoE and Weather-aware Router (ours). }
  \label{fig:hist}
\end{figure}

\begin{figure}[!t]
  \centering
  \includegraphics[width=1\linewidth]{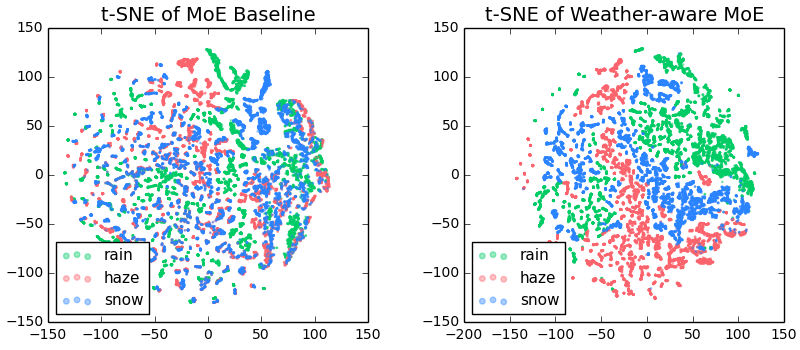}
  \caption{Comparison of t-SNE of routing weights between MoE and Weather-aware Router (ours). }
  \label{fig:tsne}
\end{figure}

\subsection{Baseline: Mixture-of-Experts Module in ViT}

We first introduce our baseline, MoE in ViT, and then analyze its problems for blind weather removal tasks.

MoE baseline~\cite{lepikhin2020gshard, fedus2021switch} includes two parts. The experts consist of multiple parallel FFNs. The input tokens are passed through each expert and then fusion by weighted summation. Weights are generated by the router. It takes each token as input and outputs the probability of each token belonging to a specific expert. We choose Top-K experts for every token. Overall, the formulation of the naive MoE module can be summarized as follows:
\begin{align}
    g (\textbf{y}^{\ell}) & = \text{Softmax} (\text{Top-K} (\textbf{W} \textbf{y}^{\ell})) \\
    \textbf{z}^{\ell +1} & = [\sum_{i=1}^{M}g_i (\textbf{y}^{\ell})\times \text{FFN}_{i} (\text{LN} (\textbf{y}^{\ell})) ] + \textbf{y}^{\ell}    
\end{align}
where $\textbf{y}^{\ell} \in \mathbb{R}^{N\times D}$ is the output of self-attention module, $\textbf{z}^{\ell} \in \mathbb{R}^{N\times D}$ is the input tokens of $\ell$-th Transformer block, $g (\textbf{z}^{\ell}) \in \mathbb{R}^{M}$ is output weights of the router, $g_i (\textbf{z}^{\ell})$ and $\text{FFN}_{i}$ are the fusion weights and the $i$-th expert respectively, and $M$ is the number of the task experts.


\subsection{Weather-aware Router}

In ideal conditions, experts will form different groups automatically with each one proficient in dealing with specific weather conditions with the help of a router with dynamic tokens routing mechanism.

To verify the hypothesis, we calculate the routing scores $s_{i, w}$ by averaging the output weights of the router for each token from images with different weather types, where $i$ is the ID of experts, $w$ is the weather type including rain, haze, and snow. To compare $s_{i, w}$ of different weather more clearly, we present the histogram of normalized routing scores $s^{norm}_{i, w} = s_{i, w} - \Sigma_{w} s_{i, w}$ in Fig.\ref{fig:hist} (a). If the hypothesis is right, the variance of $s^{norm}_{i, w}$ will be large.
However, we observe normalized routing score almost tends to zero for all weather, which means there is no preference for the router to deal with different weather. We also visualize the t-SNE of the router's weights for different weather in Fig.\ref{fig:tsne} (a), which also matches the above conclusion because the t-SNE features are mixed together.

So we rethink the router designed in naive MoE. We find in fact it's hard for the original router to select tokens according to weather conditions. The essence of the routing mechanism is to complete the pattern matching and clustering of tokens for each expert. In high-level tasks, different tokens contain abstract semantic features with high similarity~\cite{riquelme2021cv-moe-vit}, resulting in easier clustering. But in the blind weather removal task, the content and weather feature of patch embedding are coupled, making router difficult to select specific experts according to the content or weather information, which limits the assignment flexibility.

We analyze this is because the original router is hard to select tokens according to weather conditions due to the coupled content and weather features (details in supplementary material). In view of the current router's drawback, we propose a Weather-aware Router to explicitly make use of the weather feature. 
In the training stage, we employ Weather Guidance Fine-grained Contrastive Learning to learn token-level weather features as the weather representation learning branch, which will be introduced in the next section. 

These token-level weather embeddings could be used as a supplementary assignment basis for the router in the restoration branch. We concatenate weather tokens $\textbf{z}^{l}$ and content tokens $\textbf{y}^{\ell}$ alone the channel dimension to obtain $[\textbf{y}^{\ell}, \textbf{z}^{\ell}] \in \mathbb{R}^{N \times 2D}$, and then utilize a nonlinear adaptor to aggregate the feature. The adaptor's outputs are the final inputs of the router. The formulation can be summarized as follows:

\begin{equation}
    \label{weather_aware_router}
    g (\textbf{y}^{\ell}, \textbf{z}^L) = \text{Softmax} (\text{Top-K} ( \text{Adaptor} ([\textbf{y}^{\ell}, \textbf{z}^{\ell}]) \textbf{W}))
\end{equation}

In this way, WEAR can select experts based on decoupled content and weather features, which is more flexible and leads to better performance. For WEAR, the normalized routing score has a larger variance and t-SNE features are more separated, shown in Fig. \ref{fig:hist} and \ref{fig:tsne} (b) respectfully, demonstrating WEAR's effectiveness.

\subsection{Weather Guidance Fine-grained Contrastive Learning}

Contrastive Learning (CL) has been explored for learning representation to modulate networks in low-level vision tasks. DASR~\cite{wang2021dasr} proposes unsupervised degradation contrastive learning (UDCL) for blind super-resolution (blind SR). UDCL follows a prior assumption, the low-resolution degradation is the same for the same image, and different for different images, resulting in corresponding instance-level positive and negative samples. UDCL first randomly selects B low-resolution images and randomly crops two patches from each image, with patch embeddings $p_i^1, p_i^2, i=1,\dots B$. UDCL builds a online queue~\cite{he2020moco} to store negative samples $p_{queue}^j, j=1,\dots,N_{queue}$. The formulation is as follows:
\begin{equation}
    \label{udcl}
    L_{udcl} = \sum_{i=1}^{B}-\text{log}\frac{\text{exp} (p^1_{i}\cdot p^2_{i} /\tau)}{\sum_{j=1}^{N_{queue}} \text{exp} (p^1_{i}\cdot p^j_{queue}/\tau)}
\end{equation}
where $\tau \in \mathbb{R}^{+}$ is a scalar temperature parameter.

Different from blind SR, the degradation in blind weather removal shows obvious group characteristics. A real weather image consists of some basic weather elements like rain streaks, raindrops, haze, and snowflows. If we apply UDCL in blind weather, it's easy to get false negative examples. For example, patch embeddings from a rain image will be pushed to those from another rain image. 

\begin{figure}[!t]
  \centering
  \includegraphics[width=1\linewidth]{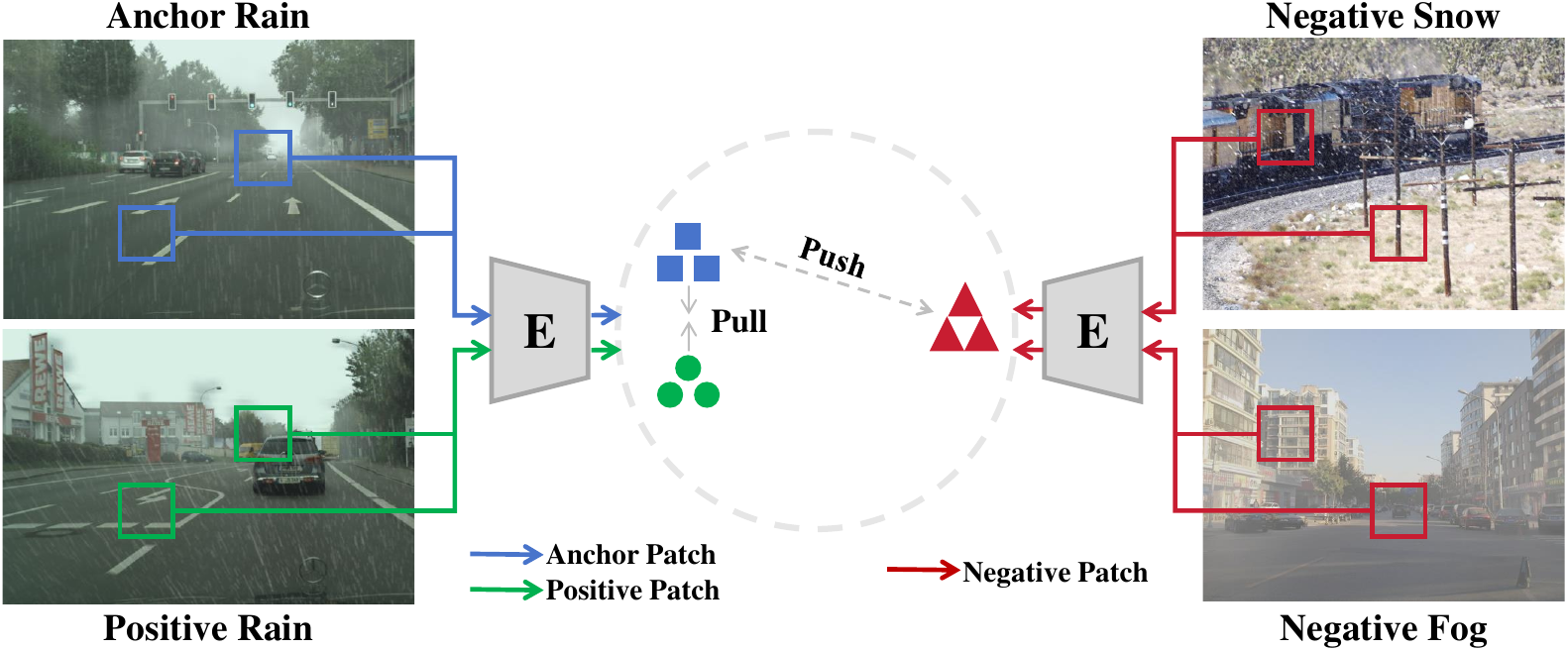}
  \caption{The illustration of proposed Weather Guidance Fine-grained Contrastive Learning. Patch embeddings from images with the same weather are regarded as positive samples and from different weather are negative samples.}
  \label{fig:WGF-CL}
\end{figure}

To better enable CL to learn blind weather features, motivated by Supervised Contrastive Learning (SCL)~\cite{khosla2020scl}, which uses classification labels to avoid false negative samples, we propose Weather Guidance Fine-grained Contrastive Learning (WGF-CL) (Fig. \ref{fig:WGF-CL}).
Though we can't obtain the weather type in test time, we can still utilize weather labels to help model training. SCL regards different augmented views (global features) of images with the same semantic label as positive samples, while WGF-CL regards all patch embeddings (local features) of images with the same weather as positive samples. The formulation is summarized as follows.

\begin{figure}[!t]
  \centering
  \includegraphics[width=1\linewidth]{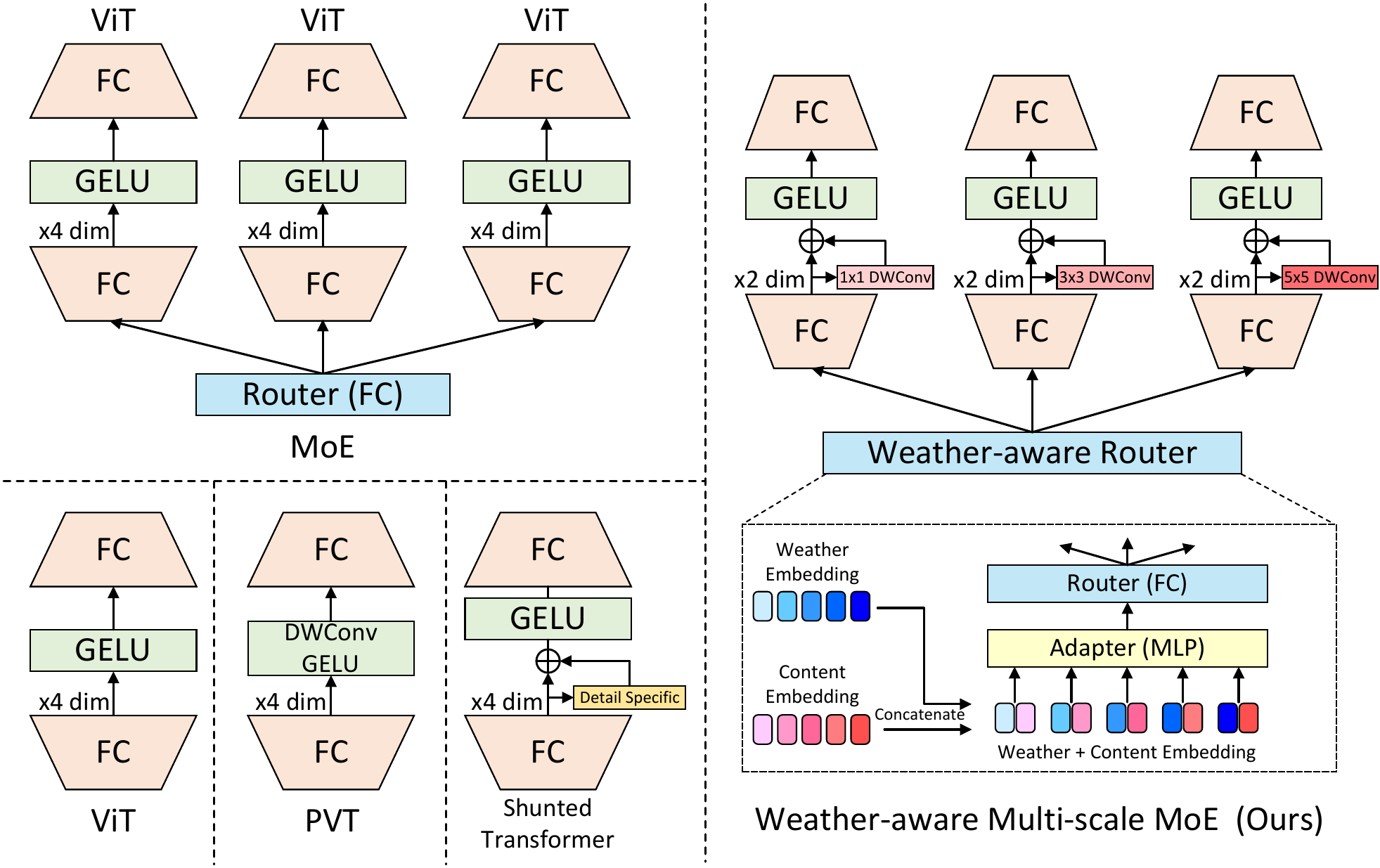}
  \caption{Comparison of ViT~\cite{dosovitskiy2020vit}, PVT~\cite{wang2021pvt}, Shunted Transformer~\cite{ren2022shunted}, MoE~\cite{lepikhin2020gshard} and proposed Weather-aware Multi-scale MoE in terms of FFN module.}
  \label{fig:Weather-aware Multi-scale MoE}
\end{figure}

\begin{align}
    \label{wgf-cl}
    \textbf{z}^{L}_{i} & = \text{ViT} (\textbf{I}_{\rm adverse}) = [z^L_{i, 1}, \dots, z^L_{i, j}, \dots, z^L_{i, N}] \\
    L_{wgf-cl} & = \sum_{i\in I} \frac{-1}{|P (i)|}\sum_{p\in P (i)}\text{log}\frac{\text{exp} (\textbf{z}^{L}_{i}\cdot \textbf{z}^{L}_{p} /\tau)}{\sum_{a\in A (i)}^{ } \text{exp} (\textbf{z}^{L}_{i}\cdot \textbf{z}^{L}_{a}/\tau)}
\end{align}
where $\textbf{z}_{i}^{L} \in \mathbb{R}^{N\times D}$ is the output of the weather feature encoder, $i\in I\equiv\{1, \dots, B\}$ is the index of minibatch, $A (i)\equiv I \backslash \{i\}$, $P (i)\equiv \{p\in A (i): y_p=y_i\}$ is the positive samples group of the $i$-th anchor sample. 

Compared to UDCL, WGF-CL utilizes weather prior information to guide contrastive learning to select appropriate positive samples and help to capture group-level weather representation.

\begin{figure*}[!t]
  \centering
  \includegraphics[width=0.9\linewidth]{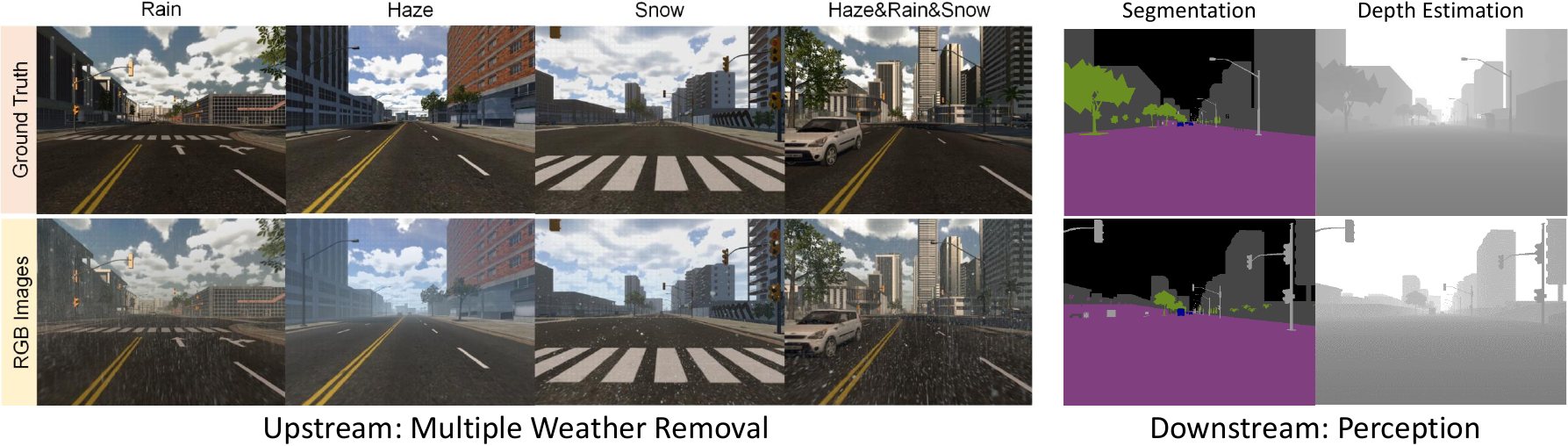}
  \caption{Examples of MAW-Sim visualizations. We provide pair of RGB images in 4 weather types. (Best view on screen)}
	\label{fig:vis_CitySpike100K}
\end{figure*}

\newcommand\MyXhlineC[0]{\Xhline{2.5\arrayrulewidth}}
\renewcommand\arraystretch{1.0}
\setlength\tabcolsep{3pt}
\begin{table*}[!t]
	\centering
	\caption{Quantitative comparison on All-Weather dataset. The GMacs and time are tested on 256$\times$256 images.}
		\begin{tabular}{c|c|c|c|c|c|c|c|c|c}
            \MyXhlineC
            \multicolumn{2}{c|}{\multirow{2}*{Type}} & \multirow{2}*{Method}  & Derain & Deraindrop  & Desnow & Average & \multirow{2}*{\makecell{Params\\(M)}} & \multirow{2}*{GMacs} & \multirow{2}*{\makecell{Time\\(ms)}} \\
            \cline{4-7}
            \multicolumn{2}{c|}{\multirow{2}{*}{}} & \multirow{2}{*}{} & 
            PSNR / SSIM  & PSNR / SSIM & PSNR / SSIM & PSNR / SSIM  & & &    \\
            \MyXhlineC
            
			\multirow{9}{*}{\begin{tabular}[c]{@{}c@{}}Task\\ Specific\end{tabular}} &\multirow{2}{*}{Derain} 
                                                   & RESCAN~\cite{li2018recurrent}    & 21.57 / 0.7255 & 24.26 / 0.8367 & 24.30 / 0.7586 & 23.38 / 0.7736 & 0.15 & 32.32 & 26.5 \\
			\multirow{11}{*}{} & \multirow{2}{*}{} & PReNet~\cite{ren2019progressive} & 23.16 / 0.8624 & 24.96 / 0.8629 & 25.19 / 0.8483 & 24.44 / 0.8579 & 0.17 & 66.58 & 23.0 \\
            \cline{2-10}
            
			\multirow{11}{*}{} & \multirow{4}{*}{Dehaze} & GridDehazeNet~\cite{liu2019griddehazenet} & 25.31  / 0.8657  & 27.32  / 0.8723  & 27.19  / 0.8457  & 26.61  / 0.8612 & 0.96 & 21.49 & 21.2  \\
			\multirow{11}{*}{} & \multirow{4}{*}{} & MSBDN-DFF~\cite{dong2020multi}            & 22.62  / 0.8217  & 21.22  / 0.8189  & 25.46  / 0.8130  & 23.10  / 0.8179 & 28.71 & 24.61 & 45.0 \\
			\multirow{11}{*}{} & \multirow{4}{*}{} & FFA-Net~\cite{qin2020ffa}                 & 27.96  / 0.8857  & 27.73  / 0.8894  & 27.21  / 0.8578  & 27.63  / 0.8776 & 4.46 & 288.34 & 55.9 \\
            \multirow{11}{*}{} & \multirow{4}{*}{} & AECR-Net~\cite{wu2021contrastive}         & 26.77  / 0.8493  & 26.54  / 0.8846  & 26.77  / 0.8509  & 26.70  / 0.8616 & 2.61 & 43.07 & 10.0  \\
            \cline{2-10}

            \multirow{11}{*}{} & \multirow{2}{*}{Desnow}  
                                                   & DesnowNet~\cite{liu2018desnownet}  & 12.73 / 0.5327 & 23.80 / 0.8440 & 21.89 / 0.7682 & 19.47 / 0.7150 & 26.15 & 107.32 & 69.2 \\
			\multirow{11}{*}{} & \multirow{2}{*}{} & HDCWNet~\cite{chen2021all}         & 14.59 / 0.6314 & 24.51 / 0.8514 & 20.21 / 0.7447 & 19.77 / 0.7425 & 6.99 & -- & -- \\
            
            \MyXhlineC
			\multicolumn{2}{c|}{\multirow{2}{*}{\begin{tabular}[c]{@{}c@{}}Task\\ Agnostic\end{tabular}}}     
                                                     & MPR~\cite{zamir2021multi}          & \subbest{28.35} / \subbest{0.9100} & \subbest{28.33} / \subbest{0.9063} & 27.77 / 0.8772 & \subbest{28.15} / \subbest{0.8978} & 3.64 & 148.55 & 63.5 \\
            \multicolumn{2}{c|}{\multirow{4}{*}{}}   & Restormer~\cite{zamir2022restormer} & 27.85 / 0.8802 & 28.32 / 0.8881 & \subbest{28.18} / \subbest{0.8684} & 28.12 / 0.8789 & 26.13 & 140.99 & 65.4 \\

            \MyXhlineC
			\multicolumn{2}{c|}{\multirow{4}{*}{\begin{tabular}[c]{@{}c@{}}Multi\\ Task in One\end{tabular}}}     
                                                     & Transweather~\cite{valanarasu2022transweather}   & 25.64 / 0.8103 & 27.37 / 0.8570 & 26.98 / 0.8305 & 26.66 / 0.8326 & 38.05 & 6.12 & 28.3 \\
            \multicolumn{2}{c|}{\multirow{3}{*}{}}   & Unified Model~\cite{chen2022learning}            & 25.81 / 0.8544 & \subbest{28.33} / 0.8832 & 27.94 / 0.8679 & 27.36 / 0.8685 & 28.71 & 24.61 & 42.2 \\
            \multicolumn{2}{c|}{\multirow{3}{*}{}}   & \textbf{WM-MoE}~(All-Weather) & \best{28.59} / \best{0.9432} & \best{29.37} / \best{0.9400} & \best{29.37} / \best{0.9014} & \best{29.11} / \best{0.9282} & 19.58 & 53.31 & 22.9 \\
            \multicolumn{2}{c|}{\multirow{3}{*}{}}   & \textbf{WM-MoE}~(Maw-Sim + Cityscapes) & --- & --- & --- & --- & 17.47 & 51.27 & 21.2 \\

            
            \MyXhlineC
            
		\end{tabular}
	\label{real_results}
\end{table*}

\subsection{Multi-scale Experts}

Different weather requires information on different receptive fields to deal with. For example, the occlusion is more serious in heavy rain, so the receptive field should be larger than in light rain for removal. However, the original MoE can only process token-level information due to point-wise FFN. 

To this end, we design Multi-scale Experts (Fig. \ref{fig:Weather-aware Multi-scale MoE}) to process information at different scales. Motivated by Inception~\cite{szegedy2015inception}, we propose grouping experts at different scales. Each expert has a parallel depth-wise convolution with $n\times n$ kernel size, denoted as $\text{DWConv}_{n}$, and different groups have different $n$, where $n \in \{1,3,5,7\}$. The formulations are as follows:

\begin{align}
    \label{multi-scale moe1}
    \textbf{z}^{\ell +1} & = [\sum_{i=1}^{M}g_i (\textbf{y}^{\ell})\times \text{FFN}_{MS, i} (\text{LN} (\textbf{y}^{\ell})) ] + \textbf{y}^{\ell}
\end{align}
\begin{align}
    \label{multi-scale moe2}
    \text{FFN}_{MS, i} (\textbf{x}) & = \text{FC} (\sigma (\text{FC} (\textbf{x}) + \text{DWConv}_{n} (\text{FC} (\textbf{x}))))
\end{align}
Where $\text{FFN}_{MS, i}$ is the $i$-th multi-scale expert, FC is the fully connected layer, and $\sigma$ is GELU. Combined with WEAR, the model can adaptively select experts with appropriate receptive fields to process different weather conditions.

\subsection{Loss Funciton}
We use the smooth-L1 loss and perceptual loss~\cite{johnson2016perceptual} for the restoration branch and weather guidance fine-grained contrastive learning (WGF-CL) loss for the weather representation learning branch. For perceptual loss, we extract features from the $3^{rd}$, $8^{th}$ and $15^{th}$ layers of VGG16~\cite{simonyan2014vgg} pretrained on ImageNet and calculate the MSE of features from the restored image and GT. For WGF-CL loss, we set $\tau=2$. The overall loss can be summarized as follows:
\begin{equation}
    L_{total} = L_{smooth-L1} + \lambda_{1} L_{perceptual} + \lambda_{2} L_{WGF-CL}
\end{equation}
where $\lambda_{1} = 0.04$ and $\lambda_{2} = 0.01$.

\section{Experiments}
In this section, we evaluate our work on MAW-Sim, All-Weather, and Cityscapes datasets and compare it with several SOTA methods. We also provide the performance comparison of downstream segmentation tasks in Cityscapes. Moreover, we implement an ablation study to demonstrate the effectiveness of each component of WM-MoE.

\subsection{Experiment Setting}
\textbf{MAW-Sim.} In order to train models for blind adverse weather removal, we collect and annotate a simulated dataset with multiple adverse weathers named \textit{MAW-Sim}. It's generated based on the Unity3D engine with a large city traffic system and a weather generation system. The images are recorded with a virtual camera on cars at 30fps. The dataset has 30 scenes. Each scene has 5 different cases, including clear day as ground truth, rain, snow, fog, and a random mix of them. Each scene has 30 frames with 1024$\times$768 resolution. We further divide the dataset for training, validation, and testing according to the ratio of 7:1:2.

\textbf{Dataset.} We evaluate WM-MoE and state-of-the-art methods on MAW-Sim, All-weather, and Cityscapes datasets.
All-weather~\cite{valanarasu2022transweather} contains data from different public datasets. The training set consists of Outdoor-Rain~\cite{li2019outdoor-rain}, Raindrop~\cite{qian2018raindrop} and Snow100K~\cite{liu2018snow100k}. The test set is sampled from Outdoor-Rain~\cite{li2019outdoor-rain}, the Raindrop test set~\cite{qian2018raindrop}, and the Snow 100k-L test set~\cite{liu2018snow100k}. Cityscapes~\cite{cordts2016cityscapes} is collected from various scenarios of outdoor street scenes in different cities. These images are recorded in normal weather conditions by video cameras mounted on cars. Foggy Cityscapes~\cite{sakaridis2018haze_cityscapes} and Rain Cityscapes~\cite{hu2019rain_cityscapes} are synthesized from the images in the Cityscapes dataset. We further combine the train sets of them for training. We also adopt the fine segmentation label of Cityscapes for downstream tasks evaluation.

\newcommand\MyXhlineG[0]{\Xhline{2.5\arrayrulewidth}}
\renewcommand\arraystretch{1.0}
\setlength\tabcolsep{6pt}
\begin{table}[!ht]
	\centering
        
	\caption{Summary of the proposed MAW-Sim datasets. We count the number of images in the train, val, and test sets of different weather types in the datasets. Mix represents the mix of the rainy, snowy, and foggy weather types. }
		\begin{tabular}{|c|c|c|c|}
            \MyXhlineG
			{MAW-Sim}            & Train  & Validation & Test \\ 
            \MyXhlineG
			Rainy      & 0.84K  & 0.12K  & 0.24K\\
			Snowy      & 0.84K  & 0.12K  & 0.24K\\
		  Foggy      & 0.84K  & 0.12K  & 0.24K\\
		  Mix      & 0.84K  & 0.12K  & 0.24K\\
		  Clear    & 0.84K  & 0.12K  & 0.24K\\
            \MyXhlineG
			Total    & 4.2K  & 0.12K  & 0.24K\\
            \MyXhlineG
		\end{tabular}
	\label{tab:Summary_datasets}
\end{table}

\newcommand\MyXhlineB[0]{\Xhline{2.5\arrayrulewidth}}
\renewcommand\arraystretch{1.0}
\setlength\tabcolsep{6pt}
\begin{table*}[!ht]
	\centering
        
	\caption{Quantitative Comparison on MAW-Sim dataset.}
		\begin{tabular}{c|c|c|c|c|c|c|c}
			\MyXhlineB
            \multicolumn{2}{c|}{\multirow{2}*{Type}} & \multirow{2}*{Method}   & Derain     & Dehaze  & Desnow & Mixed Weather & Average\\
            \cline{4-8}
            \multicolumn{2}{c|}{\multirow{2}{*}{}} & \multirow{2}{*}{} & 
            PSNR / SSIM  & PSNR / SSIM & PSNR / SSIM & PSNR / SSIM  & PSNR / SSIM      \\
            \MyXhlineB
            
			\multirow{9}{*}{\begin{tabular}[c]{@{}c@{}}Task\\ Specific\end{tabular}} &\multirow{2}{*}{Derain} 
                                                   & RESCAN~\cite{li2018recurrent}    & 21.42 / 0.7467 & 22.30 / 0.8554 & 21.58 / 0.7177 & 20.77 / 0.6539 & 21.52 / 0.7434  \\
			\multirow{11}{*}{} & \multirow{2}{*}{} & PReNet~\cite{ren2019progressive} & 28.89 / 0.9170 & 29.47 / \subbest{0.9467} & \subbest{29.24} / \subbest{0.9330} & 28.09 / \subbest{0.9081} & 28.17 / \subbest{0.9262}  \\
            \cline{2-8}
            
			\multirow{11}{*}{} & \multirow{4}{*}{Dehaze} &
            GridDehazeNet~\cite{liu2019griddehazenet}   & 28.96 / 0.8901 & 28.94 / 0.9216 & 29.22 / 0.9045 & \subbest{28.21} / 0.8759 & 28.83 / 0.8980  \\
			\multirow{11}{*}{} & \multirow{4}{*}{} & MSBDN-DFF~\cite{dong2020multi}              & 28.09 / 0.8985 & 28.55 / 0.9321 & 28.30 / 0.9118 & 27.48 / 0.8854 & 28.11 / 0.9070  \\
			\multirow{11}{*}{} & \multirow{4}{*}{} & FFA-Net~\cite{qin2020ffa}                   & 27.44 / 0.8776 & 28.73 / 0.9330 & 27.54 / 0.8810 & 26.30 / 0.8454 & 27.50 / 0.8843  \\ 
            \multirow{11}{*}{} & \multirow{4}{*}{} & AECR-Net~\cite{wu2021contrastive}           & 26.89 / 0.8584 & 27.76 / 0.9062 & 26.89 / 0.8658 & 26.29 / 0.8365 & 26.96 / 0.8667  \\ 
            \cline{2-8}

            \multirow{11}{*}{} & \multirow{2}{*}{Desnow}  
                                                   & DesnowNet~\cite{liu2018desnownet}   & 20.51 / 0.6663 & 25.01 / 0.8352 & 20.16 / 0.6198 & 18.25 / 0.5446 & 20.98 / 0.6665  \\
			\multirow{11}{*}{} & \multirow{2}{*}{} & HDCWNet~\cite{chen2021all}          & 23.98 / 0.7236 & 25.42 / 0.8392 & 24.36 / 0.8254 & 21.36 / 0.6724 & 23.78 / 0.7652  \\
            
            \MyXhlineB
			\multicolumn{2}{c|}{\multirow{2}{*}{\begin{tabular}[c]{@{}c@{}}Task\\ Agnostic\end{tabular}}}     
                                                     & MPR~\cite{zamir2021multi}            & 27.02 / 0.8881 & 28.14 / 0.9297 & 27.81 / 0.9170 & 26.39 / 0.8753 & 27.34 / 0.9025  \\
            \multicolumn{2}{c|}{\multirow{4}{*}{}}   & Restormer~\cite{zamir2022restormer}  & 28.23 / 0.9016 & 28.21 / 0.9350 & 28.30 / 0.9137 & 27.11 / 0.8859 & 27.96 / 0.9091  \\

            \MyXhlineB
			\multicolumn{2}{c|}{\multirow{3}{*}{\begin{tabular}[c]{@{}c@{}}Multi\\ Task in One\end{tabular}}}     
                                                     & Transweather~\cite{valanarasu2022transweather}   & 22.55 / 0.6608 & 22.97 / 0.7163 & 21.90 / 0.5857 & 21.38 / 0.5507 & 22.20 / 0.6284  \\
            \multicolumn{2}{c|}{\multirow{3}{*}{}}   & Unified Model~\cite{chen2022learning}            & \subbest{29.04} / \subbest{0.8913} & \subbest{29.63} / 0.9312 & 29.11 / 0.9002 & 28.00 / 0.8701 & \subbest{28.95} / 0.8982  \\
            \multicolumn{2}{c|}{\multirow{3}{*}{}}   & \textbf{WM-MoE}                                         & \best{30.38} / \best{0.9423} & \best{31.16} / \best{0.9656} & \best{30.77} / \best{0.9522} & \best{29.52} / \best{0.9328} & \best{30.46} / \best{0.9482}  \\

            
            \MyXhlineB
            
		\end{tabular}
	\label{synthetic_results}
\end{table*}

\newcommand\MyXhlineE[0]{\Xhline{2.5\arrayrulewidth}}
\renewcommand\arraystretch{1.0}
\setlength\tabcolsep{6pt}
\begin{table*}[ht]
	\centering
	\caption{Quantitative Comparison on Cityscapes. }
		\begin{tabular}{c|c|c|c|c|c|c|c|c|c|c}
            \MyXhlineE
            \multicolumn{2}{c|}{\multirow{3}*{Type}} & \multirow{3}*{Method} & \multicolumn{4}{c|}{Derain}  & \multicolumn{4}{c}{Dehaze} \\
            \cline{4-11}
            \multicolumn{2}{c|}{\multirow{2}{*}{}} & \multirow{2}{*}{} & 
            \multicolumn{2}{c|}{Upstream} & \multicolumn{2}{c|}{Downstream} & \multicolumn{2}{c|}{Upstream} & \multicolumn{2}{c}{Downstream}      \\
            \cline{4-11}
            \multicolumn{2}{c|}{\multirow{2}{*}{}} & \multirow{2}{*}{} & 
            PSNR $\uparrow$ & SSIM $\uparrow$   & mIoU $\uparrow$ & mAcc $\uparrow$ & PSNR $\uparrow$ & SSIM $\uparrow$  & mIoU $\uparrow$ & mAcc $\uparrow$      \\
            \MyXhlineE
            
			\multirow{7}{*}{\begin{tabular}[c]{@{}c@{}}Task\\ Specific\end{tabular}} &\multirow{2}{*}{Derain} 
                                                   & RESCAN~\cite{li2018recurrent}    & 19.11 & 0.9118 & 0.1007 & 0.2064      & 16.96 & 0.9033 & 0.1262 & 0.2116 \\
			\multirow{11}{*}{} & \multirow{2}{*}{} & PReNet~\cite{ren2019progressive} & 19.95 & 0.8822 & 0.1321 & 0.2943      & 18.22 & 0.8729 & 0.3305 & 0.2508 \\
            \cline{2-11}
            
			\multirow{11}{*}{} & \multirow{4}{*}{Dehaze} & GridDehazeNet~\cite{liu2019griddehazenet} & 22.08 & 0.9171 & 0.4259 & 0.6945    & 23.18 & 0.9183 & 0.4453 & 0.7108 \\
			\multirow{11}{*}{} & \multirow{4}{*}{} & MSBDN-DFF~\cite{dong2020multi}            & 26.26 & 0.8853 & 0.1744 & 0.2733    & 26.79 & 0.8903 & 0.3426 & 0.4019 \\
			\multirow{11}{*}{} & \multirow{4}{*}{} & FFA-Net~\cite{qin2020ffa}                 & 28.29 & 0.9411 & 0.3458 & 0.5799    & 28.96 & 0.9432 & 0.4257 & 0.6005 \\
            \multirow{11}{*}{} & \multirow{4}{*}{} & AECR-Net~\cite{wu2021contrastive}         & 26.27 & 0.9075 & 0.2230 & 0.3516    & 27.75 & 0.9062 & 0.3714 & 0.4652 \\
            \cline{2-11}

            \MyXhlineE
			\multicolumn{2}{c|}{\multirow{2}{*}{\begin{tabular}[c]{@{}c@{}}Task\\ Agnostic\end{tabular}}}     
                                                     & MPR~\cite{zamir2021multi}           & \subbest{32.68} & \best{0.9810} & \subbest{0.4657} & \subbest{0.7580}     & \subbest{29.73} & \best{0.9752} & \subbest{0.4537} & \subbest{0.7301} \\
            \multicolumn{2}{c|}{\multirow{4}{*}{}}   & Restormer~\cite{zamir2022restormer} & 28.06 & 0.9630 & 0.4383 & 0.6833     & 22.72 & 0.9411 & 0.4085 & 0.6920 \\

            \MyXhlineE
			\multicolumn{2}{c|}{\multirow{3}{*}{\begin{tabular}[c]{@{}c@{}}Multi\\ Task in One\end{tabular}}}     
                                                     & Transweather~\cite{valanarasu2022transweather}   & 24.08 & 0.8481 & 0.4425 & 0.671      & 22.56 & 0.8736 & 0.3643 & 0.6105 \\
            \multicolumn{2}{c|}{\multirow{3}{*}{}}   & Unified Model~\cite{chen2022learning}            & 28.25 & 0.9504 & 0.4190 & 0.7246      & 27.96 & 0.9167 & 0.4336 & 0.7231 \\
            \multicolumn{2}{c|}{\multirow{3}{*}{}}   & \textbf{WM-MoE}           & \best{32.99} & \subbest{0.9755} & \best{0.4686} & \best{0.7701} & \best{31.31} & \subbest{0.9647} & \best{0.4545} & \best{0.7473} \\

            
            \MyXhlineE
            \multicolumn{3}{c|}{Lower Bound}  & \textbackslash{} & \textbackslash{} & 0.3416 & 0.6020 & \textbackslash{} & \textbackslash{}     & 0.4250 & 0.6391 \\
            \MyXhlineE
            \multicolumn{3}{c|}{Upper Bound}             & \textbackslash{} & \textbackslash{} & 0.4594 & 0.7685 & \textbackslash{} & \textbackslash{}     & 0.4594 & 0.7685 \\
            \MyXhlineE
            
		\end{tabular}
	\label{cityscapes_results}
\end{table*} 

\textbf{Implementation Details.}
WM-MoE is trained for 200 epochs with 32 batch sizes. We adopt the AdamW optimizer and Cosine scheduler with initial learning rates $2e-4$ gradually reduced to $2e-6$ and the warm-up strategy for 3 epochs. We randomly crop images to 256 × 256 for training and apply a non-overlap crop for the same patch size in the test. For data augmentation, we use random flips and rotations. 

\begin{figure*}[ht]
  \centering
  \includegraphics[width=0.92\linewidth]{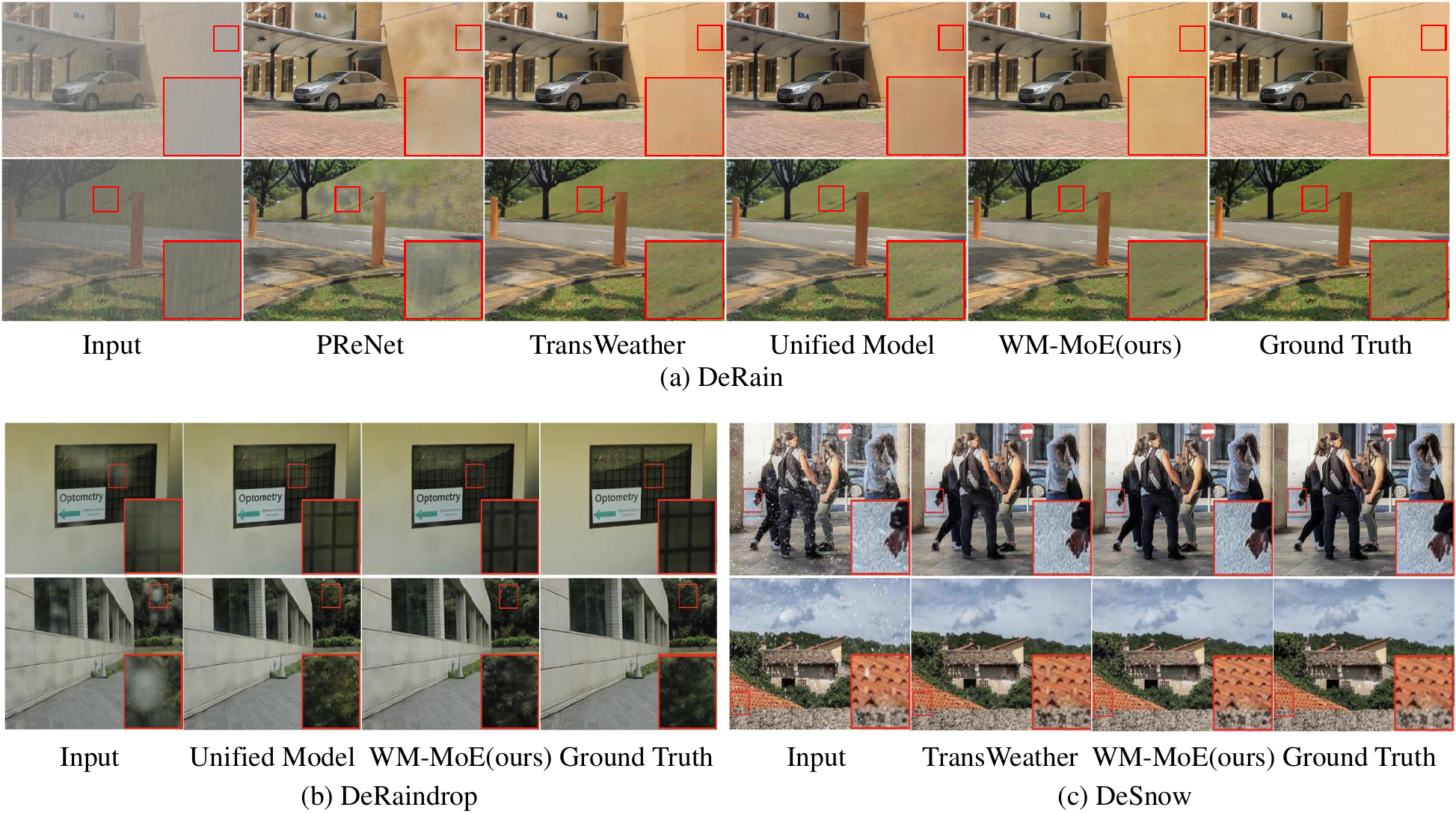}
  \caption{Qualitative results to visualize the deweather performance on All-Weather.}
  \label{fig:vis_allweather}
\end{figure*}

\begin{figure*}[!t]
	\centering
	\includegraphics[width=0.95\linewidth]{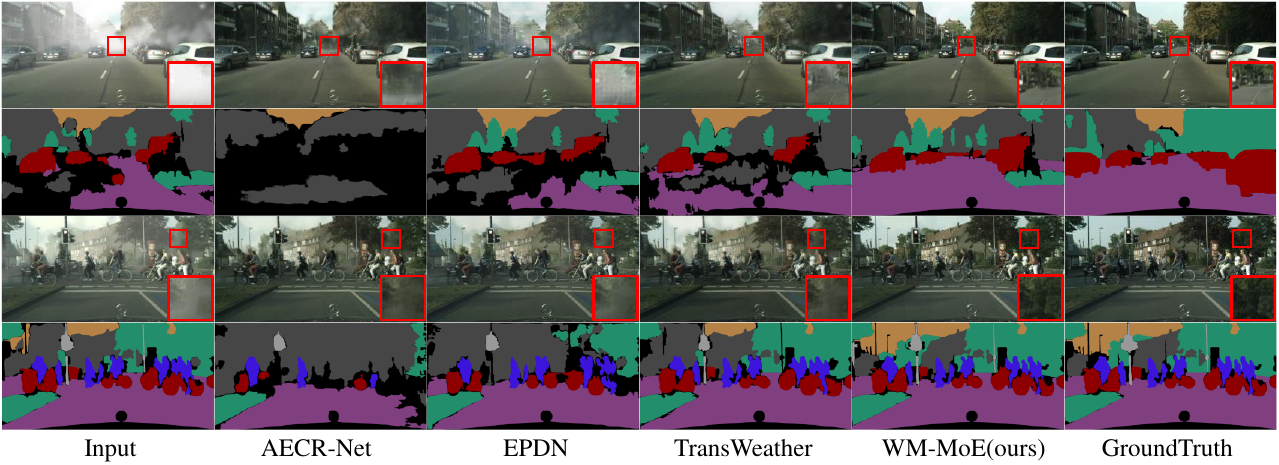}
	\caption{Qualitative results of the deweather performance and downstream semantic segmentation on Cityscapes.}
	\label{fig:vis_cityscapes}
\end{figure*}

\begin{figure*}[ht]
  \centering
  \includegraphics[width=0.8\linewidth]{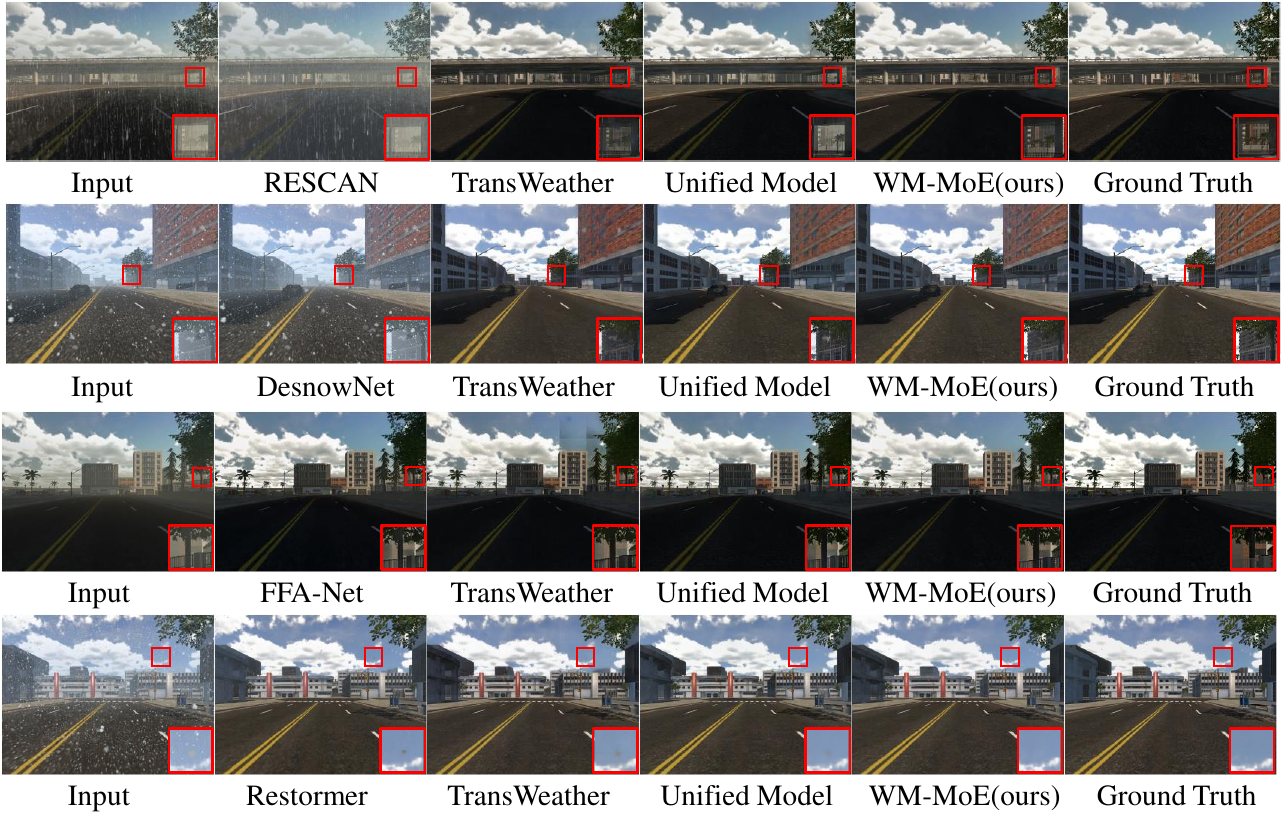}
  \caption{Qualitative results to visualize the deweather performance on MAW-Sim.}
  \label{fig:vis_mawsim_upstream}
\end{figure*}

\subsection{Network Details} \label{network details}
\textbf{Task-shared Head.}
We adopt a similar head setting as IPT~\cite{chen2021ipt}, which consists of one convolution layer and two residual blocks. The former parameters are $3 \times 3$ kernel size, $3$ input channels, and $32$ output channels. The latter block consists of a shortcut and parallel two convolution layers with $3 \times 3$ kernel size, $32$ input channels, and $32$ output channels. The spatial resolution of feature maps remains unchanged in the process.

\textbf{Transformer Encoder.}
We first obtain the patch embedding from feature maps by a non-overlap convolution projection and flatten operation. We set the dimension $D$ to $256$, $384$ and $256$ for MAW-Sim, Allweather~\cite{valanarasu2022transweather} and Cityscapes~\cite{hu2019rain_cityscapes, sakaridis2018haze_cityscapes} respectfully. The resultant sequence $[ B, N, D]$ is the input of Transformer, where $B$ is the batch size, $N$ is the number of patches and $D$ is the embedding dimension.

For the restoration branch, the number of Transformer blocks is $2$. In each block, standard multi-head self-attention module with $8$ heads is adopted. The weather-aware token-level router consists of one linear layer and softmax function with $2 \times D$ and $E$ as input and output channels, where $E$ denotes the number of experts. Before getting into the router, we adopt an adaptor with a two-layer MLP to process the concatenation of the content and weather token embeddings. $E$ multi-scale experts are employed and divided into 4 groups, with $1\times 1$, $3\times 3$, $5\times5 $ and $7\times 7$ for their bypass depth-wise convolution layer respectfully. To reduce the parameters and flops, the hidden layer dimension of each expert is set to $2D$ instead of commonly used $4D$~\cite{dosovitskiy2020vit, fedus2021switch}.

For the weather representation captured branch, we use ViT~\cite{dosovitskiy2020vit} with $2$ blocks, including multi-head self-attention module, and feed-forward network. All settings are the same as the restoration branch. 

\newcommand\MyXhlineD[0]{\Xhline{2.5\arrayrulewidth}}
\renewcommand\arraystretch{1}
\setlength\tabcolsep{3pt}
\begin{table*}[!t]
    \centering
    \caption{Ablation studies on WM-MoE. n is the number of experts and k it the TopK gate.}
    \begin{tabular}{cccc|c|c|c|c|c|c|c|c}
    \MyXhlineD
    \multirow{2}{*}{Baseline} & 
    \multirow{2}{*}{Moe-Token} & 
    \multirow{2}{*}{\makecell[c]{Weather-aware \\ router}} & 
    \multirow{2}{*}{\makecell[c]{Multi-scale \\ experts}} & 
    \multicolumn{2}{c|}{Parameters (M)} &
    \multicolumn{2}{c|}{Flops(G)} &
    \multicolumn{2}{c|}{PSNR} & 
    \multicolumn{2}{c}{SSIM} \\
    \cline{5-12}
    & & & & n4-k4 & n16-k4 & n4-k4 & n16-k4& n4-k4 & n16-k4 & n4-k4 & n16-k4
    \\
    \MyXhlineD
    \ding{51} &           &           &           & \multicolumn{2}{c|}{4.95} & \multicolumn{2}{c|}{71.96} & \multicolumn{2}{c|}{29.51} & \multicolumn{2}{c}{0.9365} \\ \cline{5-12}
    \ding{51} & \ding{51} &           &           & 6.53 & 12.86 & 75.20 & 88.12 & 29.69 & 29.84 & 0.9400 & 0.9420 \\ \cline{5-12}
    \ding{51} & \ding{51} & \ding{51} &           & 10.78 & 17.11 & 88.88 & 101.82 & 29.80 & 29.99 & 0.9406 & 0.9439 \\ \cline{5-12}
    \ding{51} & \ding{51} & \ding{51} & \ding{51} & 10.87 & 17.47 & 89.06 & 102.54 & 30.18 & 30.33 & 0.9441 & 0.9467 \\ \cline{5-12}
    
    \MyXhlineD
    \end{tabular}
    \label{tab:ablation}
\end{table*}

\newcommand\MyXhlineF[0]{\Xhline{2.5\arrayrulewidth}}
\renewcommand\arraystretch{1.1}
\setlength\tabcolsep{8pt}
\begin{table*}[!ht]
    \centering
    \caption{Ablation study on the proposed Weather-aware Multi-scale Mixture-of-Experts Module in the MAWSim dataset, including the number of experts and the TopK router. K=1 \& N=1 represents our baseline without MoE Module}
    \begin{tabular}{c|c|c|c|c|c|c|c|c|c|c}
    \MyXhlineF
    \multirow{2}{*}{K / N} & 
    \multicolumn{2}{c|}{1} & \multicolumn{2}{c|}{4} & \multicolumn{2}{c|}{8} & \multicolumn{2}{c|}{16} & \multicolumn{2}{c|}{32} \\
    \cline{2-11}
    & PSNR & SSIM & PSNR & SSIM & PSNR & SSIM & PSNR & SSIM & PSNR & SSIM\\
    \MyXhlineF
    1     & \multicolumn{1}{c|}{29.51} & \multicolumn{1}{c|}{0.9365} & 29.65 & 0.9387 & 29.60 & 0.9383 & 29.63 & 0.9387 & 29.70 & 0.9395 \\
    2     & \textbackslash{} & \textbackslash{} & 29.98 & 0.9427 & 30.04  & 0.9440 & 30.03 & 0.9449 & 30.12 & 0.9445 \\
    3     & \textbackslash{} & \textbackslash{} & 29.91 & 0.9423 & 30.09 & 0.9446 & 30.25 & 0.9462 & 30.30 & 0.9461 \\
    4     & \textbackslash{} & \textbackslash{} & 30.09 & 0.9439 & 30.24 & 0.9454 & 30.29 & 0.9468 & 30.31 & 0.9467 \\
    \MyXhlineF
    \end{tabular}
    \label{tab:ablationKN}
\end{table*}

\newcommand\MyXhilneF[0]{\Xhline{2.5\arrayrulewidth}}
\renewcommand\arraystretch{1}
\setlength\tabcolsep{3pt}
\begin{table}[!t]
    \centering
    \caption{Ablation studies on representation learning.}
    \begin{tabular}{c|c|c|c|c}
    \MyXhilneF
    \multirow{2}{*}{\makecell{Representation\\Learning Method}} & 
    \multicolumn{2}{c|}{PSNR} & 
    \multicolumn{2}{c}{SSIM} \\
    \cline{2-5}
    & n4-k4 & n16-k4 & n4-k4 & n16-k4
    \\
    \MyXhilneF
    Classification & 30.10 & 30.28 & 0.9439 & 0.9466 \\ \hline
    UDCL & 29.98 & 30.25 & 0.9436 & 0.9468 \\ \hline
    WGF-CL & 30.18 & 30.33 & 0.9441 & 0.9467 \\
    \MyXhilneF
    \end{tabular}
    \label{tab:ablation2}
\end{table}

\textbf{Task-shared Tail.}
The task-shared tail consists of $4$ sequential blocks to adjust the channels of feature maps reconstructed from the Transformer output. The residual block and convolution layer are used twice alternately. The input channels are $128$ and the output channels of each layer are $128, 64, 64, 3$ respectively. $3 \times 3$ kernel size is adopted for all of them. In the end, We utilize the Sigmoid function to normalize the final output in the range of 0 to 1.

\subsection{Comparison with State-of-the-art Methods} \label{formal version}

\textbf{Baselines.} We compare our WM-MoE with task-specific, task-agnostic, and multi-task-in-one methods. Task-specific group includes derain (RESCAN~\cite{li2018recurrent} and PReNet~\cite{ren2019progressive}), dehaze (GridDehazeNet~\cite{liu2019griddehazenet}, MSBDN-DFF~\cite{dong2020multi}, FFA-Net~\cite{qin2020ffa}, and AECR-Net~\cite{wu2021contrastive}), and desnow (DesnowNet~\cite{liu2018desnownet} and HDCWNet~\cite{chen2021all}). Task-agnostic group includes MPR~\cite{zamir2021multi}, and Restormer~\cite{zamir2022restormer}. The multi-task-in-one group includes Transweather~\cite{valanarasu2022transweather} and Unified Model~\cite{chen2022learning}. We retrain the models using open-source code. For downstream tasks, we implement experiments of the lower and upper bound, which take weather and clean images as downstream input respectively. 

\textbf{Metrics.} For the quantitative evaluation of blind weather removal, we adopt PSNR and SSIM as metrics. We calculate the metrics in RGB space. For downstream semantic segmentation tasks, we take mIoU, and mAcc as our metrics.

\begin{figure*}[!t]
  \centering
  \includegraphics[width=0.65\linewidth]{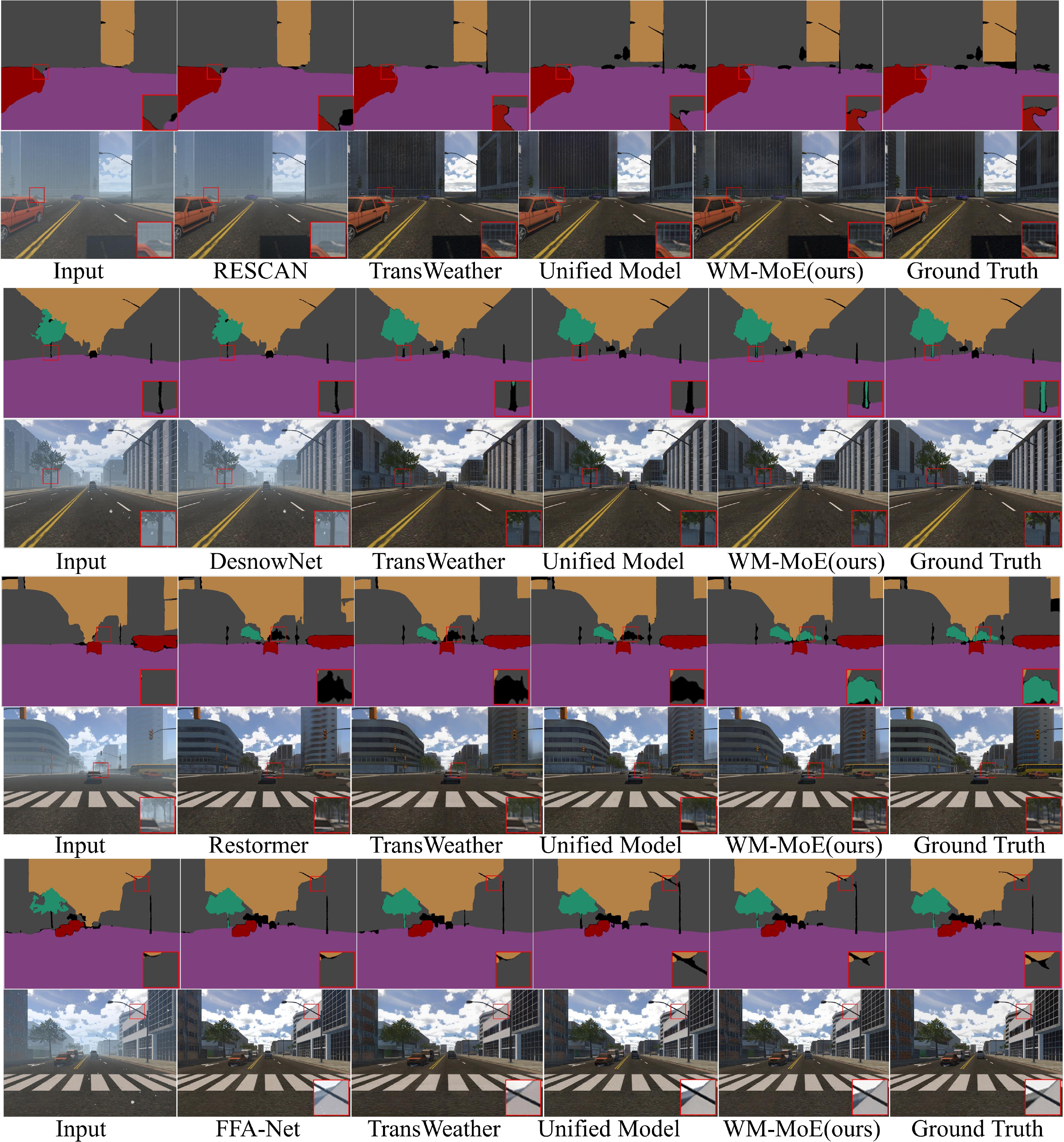}
  \caption{Qualitative results to visualize the downstream semantic segmentation performance on MAW-Sim.}
  \label{fig:vis_mawsim_downstream}
\end{figure*}

\subsubsection{Quantitative Comparison and Analysis} \ \ \ \ \

\textbf{Main results.} We report several weather and the average results in Tables \ref{real_results}, \ref{synthetic_results}, and \ref{cityscapes_results}. It can be noted that our method achieves SOTA performances on both MAW-Sim and All-weather. WM-MoE also obtains competitive results on Cityscapes. For task-specific methods designed for specific deweather tasks, WM-MoE outperforms all of them by a significant margin. Compared with task-agnostic methods, our method also achieves more promising results, especially in mixed weather. In addition, our method still performs better than another multi-task in one work, regardless of scenario and weather conditions. Thanks to the Weather-aware Router and Multi-scale Experts, the model enhances the capacity to process complex weather conditions and exploit multi-scale features, which handles blind weather better.

\textbf{Downstream results.} As for downstream results in Tables \ref{cityscapes_results}, it can be seen that weather removal benefits segmentation tasks. WM-MoE achieves competitive results on deweather tasks while helping the downstream model obtain promising results, which is close to the upper bound performance.

\subsection{Ablation Study}

  
        
        

 \textbf{Ablations on proposed modules}. 
We evaluate the contribution of each component to our method (Table~\ref{tab:ablation}). The baseline is a vision Transformer with task-shared convolution head and tail. First, we replace the original FFN with naive MoE. Then we verify the WEAR to replace the linear layer router and the ME to replace point-wise FFN experts, respectively. Here we present two parameters of the number of experts $n$ and TopK gate $k$. More ablations about $n$ and $k$ are in the supplementary material. All the ablations are conducted in MAW-Sim. The training settings are the same as Table~\ref{synthetic_results}. 
It's been shown that every design in WM-MoE could lead to a performance gain.  

\textbf{Ablations on representation learning methods.}
We also compare different representation learning methods (Table~\ref{tab:ablation2}). WGF-CL achieves the best performance. It's worth noting that the result of UDCL drops obviously, proving the hypothesis that instance-level CL against group characteristics in blind weather results in false negative samples.

\begin{figure}[!ht]
  \centering
  \includegraphics[width=1.0\linewidth]{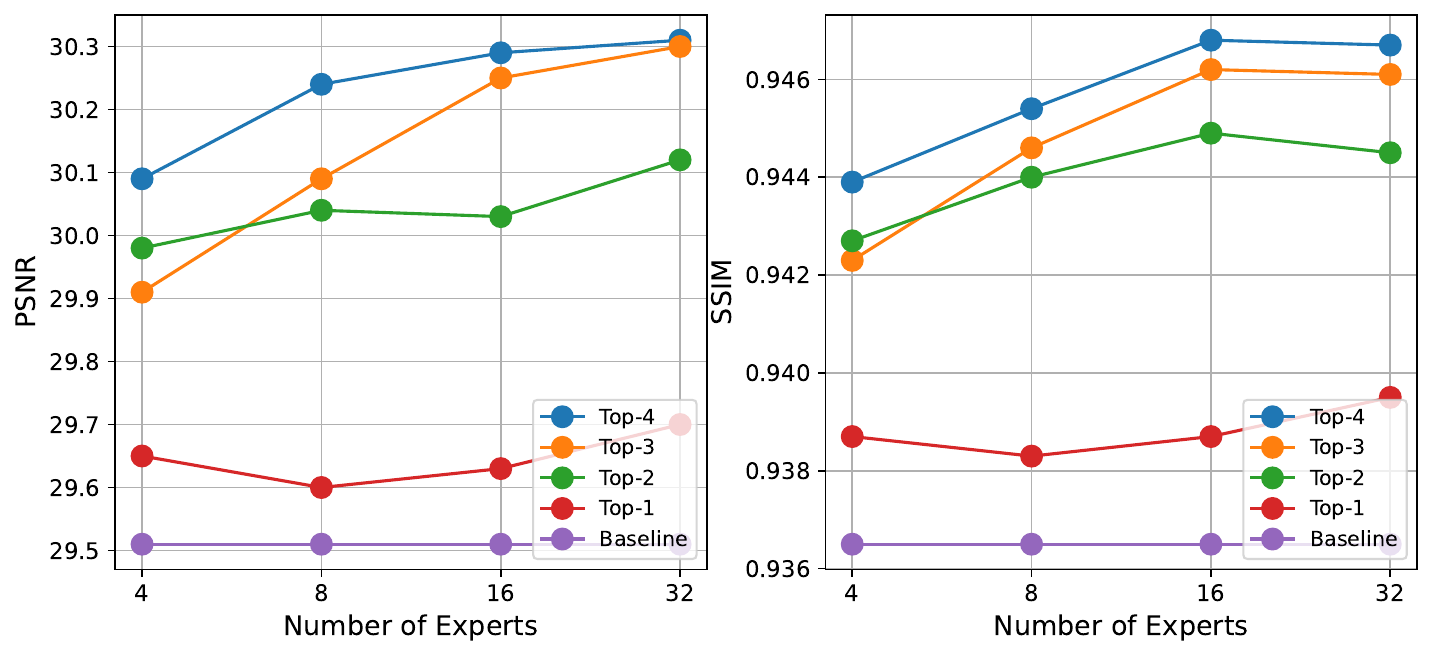}
  \caption{Ablation study on the proposed Weather-aware Multi-scale Mixture-of-Experts Module. The same results with Table \ref{tab:ablationKN}}
  \label{fig:moe_ablation}
\end{figure}

\textbf{Ablations on the number of experts and top-k gate.}
We present more ablation experiments on different N \& K in the MAW-Sim dataset with the same training setting as the main experiments. The concrete value of PSNR and SSIM is in Table \ref{tab:ablationKN} and the corresponding line chart is in Figure \ref{fig:moe_ablation}.

From the Table \ref{tab:ablationKN} and Figure \ref{fig:moe_ablation}, we can summarize the following conclusion:

 (1) The proposed MoE module can improve the baseline consistently, whatever N or K is.

 (2) When K is fixed, the performance is guaranteed to go up with N increasing from 4 to 32, and the PSNR-N growth curve is close to a logistics curve like the formulation of PSNR, which demonstrates the scaling ability of MoE. Due to limited computing resources, we don't explore more experts, but we will take it as the direction for future improvements.

 (3) For K=1, the performance increase unsteadily. We think this is because smaller K and larger N make each expert get fewer tokens to obtain strong capacity when the trainset and training time are limited.

 (3) When N is fixed, larger K brings about better performance while more computation at the same time because K controls the sparsity of MoE module~\cite{fedus2021switch, shazeer2017lstm-moe}.

After considering the overall performance and efficiency, we choose N = 16 \& K = 4 as the final version. 

\section{Conclusion}
In this work, we study the blind weather removal problem. To better release the potential of MoE in the setting, we propose a novel Weather-aware Multi-scale MoE (WM-MoE), which consists of the Weather-aware Router (WEAR) to assign the correct weather experts to tokens and Multi-scale Experts (ME) to improve the spatial modeling capability. We also propose Weather Guidance Fine-grained Contrastive Learning (WGF-CL) to decouple the weather and content information from the input image. Our method achieves SOTA performance in MAW-Sim, All-Weather, and competitive results in Cityscapes. Experiments also demonstrate that WM-MoE is beneficial to the downstream segmentation task.

\bibliographystyle{IEEEtran}
\bibliography{WM-MoE}


 




\vfill

\end{document}